\documentclass{article}

\usepackage{PRIMEarxiv}
\usepackage{amsmath}
\usepackage{authblk}
\usepackage[utf8]{inputenc} % allow utf-8 input
\usepackage[T1]{fontenc}    % use 8-bit T1 fonts
\usepackage{hyperref}       % hyperlinks
\usepackage{url}            % simple URL typesetting
\usepackage{booktabs}       % professional-quality tables
\usepackage{amsfonts}       % blackboard math symbols
\usepackage{nicefrac}       % compact symbols for 1/2, etc.
\usepackage{microtype}      % microtypography
\usepackage{lipsum}
\usepackage{fancyhdr}       % header
\usepackage{graphicx}       % graphics
\graphicspath{{media/}}     % organize your images and other figures under media/ folder

\title{Training like Playing: A Reinforcement Learning And Knowledge Graph-based framework for building Automatic Consultation System in Medical Field
\thanks{\textit{\underline{Corresponding author}}: 
\textbf{Keke Tang \  email: \ tkk2012@gmail.com}} 
}

\author[\space\space 1,3]{Yining Huang \thanks{huangyining1987@gmail.com}}
\author[\space\space1]{Meilian Chen \thanks{523062863@qq.com}}
\author[\space\space1,2]{*Keke Tang  \thanks{tkk2012@gmail.com}}
\affil[1]{Sinohealth research institution}
\affil[2]{Shenyang institute of computing technology, Chinese academy of sciences}
\affil[3]{School of Politics and Public Administration, South China Normal University}

\begin{document}
\maketitle

\begin{abstract}
We introduce a framework for AI-based medical consultation system with knowledge graph embedding and reinforcement learning components and its implement. Our implement of this framework leverages knowledge organized as a graph to have diagnosis according to evidence collected from patients recurrently and dynamically. According to experiment we designed for evaluating its performance, it archives a good result. More importantly, for getting better performance, researchers can implement it on this framework based on their innovative ideas, well designed experiments and even clinical trials.
\end{abstract}

% keywords can be removed
% \keywords{First keyword \and Second keyword \and More}

\section*{Introduction}
In recent years, medical resources in China have reached a state of in short supply. In particular, doctors with rich experience in the tertiary hospitals in big cities have to face a large number of patients coming to visit every day, but most of them do not need to go to the big hospitals for treatment. In addition, because most ordinary people have a relatively shallow understanding of medical knowledge, the phenomenon of indiscriminate medical treatment occurs, which causes a certain degree of waste of time and medical resources. These phenomena reflect two problems. The first one is the demand side. With the improvement of residents’ living standards and income levels, residents’ willingness to consume medical services is also increasing. In other words, more and more people pay attention to health, but most of them lack of professional guidance and support. The second one is the supply side, this phenomenon also shows the low level of primary medical institutions and the low level of residents’ trust in primary medical services. In recent years, the country has also actively promoted hierarchical diagnosis and treatment to avoid waste of medical resources. One of the solutions is to promote light consultation, which means without consulting in person but in the form of online style or robotic question answering. Our proposal is an auxiliary intelligent consultation system based on the knowledge graph embedding of the translation method, deep neural network and Policy Gradient based reinforcement learning, and mainly implements the demand of light consultation by man-machine automatic interaction.

Knowledge graph\cite{hogan2020knowledge} is one of the technologies on which our proposal relies. The knowledge graph is a way of organizing knowledge. It is composed of multiple triplets. The triplets consist of head entity nodes, relations, and tail entity nodes, which are used to represent facts. Knowledge graph technology is an interdisciplinary technology, composed of graph theory\cite{biggs1986graph}, computer science\cite{brookshear1991computer}, information visualization technology\cite{ware2019information} and so on. It has been developing for many years, and the research and development heat in recent years has a continuous upward trend\cite{ji2021survey}. Nevertheless, it can be applied in many aspects, including knowledge organization, visualization, information retrieval, question answering system, recommendation system etc.

The embedded representation\cite{wang2017knowledge} of knowledge graphs has gained more and more attention in recent years. Symbolic reasoning\cite{chen2020review} and deep learning\cite{lecun2015deep} based inference are not compatible in early years. However, since the development of embedded representation methods and deep learning, it become more and more applicable\cite{liang2016neural}. In other words, that has established a bridge for the fusion of the above two technologies, and can make comprehensive use of the advantages of both sides. For symbolic reasoning, the interpretability\cite{chakraborty2017interpretability} is higher than deep learning based methods. On the other hand, deep learning based methods can have fuzzy or probability reasoning, which is lacked of by symbolic reasoning methods.

In addition, deep learning and reinforcement learning\cite{sutton2018reinforcement} theories are also developing rapidly with the help of breakthroughs in the technical bottleneck of the underlying heterogeneous computing architecture\cite{capra2020updated}. These two machine learning technologies have also seen more and more breakthroughs in both theoretical and practical aspects\cite{silver2016mastering,silver2017mastering}, which are reflected in computer vision, Natural language processing and other applications in the field of artificial intelligence\cite{krizhevsky2012imagenet,goodfellow2014generative,mikolov2010recurrent}. Deep learning enables computers to initially possess human perception, that is, the ability to extract information from raw data, including images and natural language, etc. Reinforcement learning simulates the process of human learning skills, and dynamically adjusts its own strategy for a certain goal through external feedback and rewards to maximize the effect. Deep reinforcement learning is the combination of the two technologies mentioned above, which enables the machine to combine the advantages of the two to complete some more complex tasks, such as playing games\cite{mnih2013playing} and dialog policy management\cite{shah2016interactive} etc.

However, in the medical field, there are few cases of comprehensively using the above technologies for application. The innovative point of our proposal is to integrate the various advantages of the above technologies to solve important problems in the current practical scenario. We use the knowledge graph to organize the knowledge, and then use the embedded representation to map knowledge into the vector space, such that deep learning technology can be used in the down stream processes. And at the same time, to build the intelligent consultation system, our solution uses the reinforcement learning to solve the complex problems of the actual scene. In addition, based on the design of the underlying infrastructure, including the design of system availability, fault tolerance, scalability, and operational performance, we meet the requirements of the real scene. Moreover, our proposal not only has core service functions, but also has continuous systematical stability.

\section*{Related works}
\label{sec:headings}

\subsection*{Electronic Medical Consultation System}
Electronic medical consultation systems leverage labor force of doctors and general physicians to provide high quality medical service to patients, which aims to improve the ability of practice staff to manage workload and access\cite{banks2018use}. Meanwhile, it lowers costs of health care while maintaining or improving quality of disease management and health promotion. In general, they are applied in the form of telephone, email, text and video etc. To assess the effects which electronic consultation system brings, many researches are conducted by using interviews and statistical analysis\cite{banks2018use,atherton2012email,campbell2014telephone,farr2018implementing,mold2019electronic}. \cite{edwards2017use} conducts an evaluation of an online consultation system in primary care by designing a 15-month observation study and discover that the utilization of it is very low. However, a point of view indicates that low reported use may not reflect the low interest in using electronic consultation system. They consider that local heath policies and technical infrastructures may be a potential factor behind it\cite{newhouse2015patient}. Some researches\cite{hanna2012place,brant2016using,randhawa2019exploration} aim to explore views and attitudes of General Practitioners towards the new non-face-to-face consultation technologies in routine primary care by conducting interview and statistical analysis. Generally, General Practitioners consider that the electronic consultation system might be useful and offer benefits for the practice but will increase workload and costs\cite{banks2018use,edwards2017use}. Moreover, some researches\cite{atherton2013experiences,randhawa2019exploration,mold2019electronic} point out concerns from other perspective, including privacy and security of patients’ data, safety as well as quality in terms of medical service after adoption of the system, cost and sustainability under low usage etc. Hence, some GP are willing to consider using new technologies in administrative or less complex tasks\cite{hanna2012place}. \cite{atherton2016we} concludes some points which we need to consider when planing, implementing and researching the adoption of alternatives to traditional face-to-face consultation with new form of electronic consultation in primary care, which includes that different types of users, patients and members of staff in primary care, who might be affected need to be consider, and planing and researching on new technologies should be attuned to unexpected consequences. Such that, practitioner and researchers should start with the least controversial and most promising changes for the practice. In addition, \cite{hansen2014patients} points out that different types of user, patient and general practitioner, have different requirement in email-based consultation system. This idea should be generalized to the task of building comprehensive consultation system. 

\subsection*{AI-based Implementation}
As the development of Artificial Intelligence in recent years, more and more works of building consultation systems based on related technologies emerge. One of the key components in our proposal is knowledge graph. The main responsibility of knowledge graph is organizing knowledge and providing basis for reasoning. Early in 2007, \cite{smith2007obo} describes the Open Biomedical Ontologies Foundry in which ontologies could be designed to be interoperable and logically well formed and expanded continuously on the basis of an evolving set of shared principles governing ontology development. \cite{scheuermann2009toward} outlines the coherent framework for representing entities and relations. \cite{cowell2010infectious} reviews the most widely used vocabulary resources relevant to infectious disease, and \cite{schriml2012disease} introduces a comprehensive knowledge base of inherited, developmental and acquired human disease, which could be considered as the fundamental knowledge base of implementing AI consultation system. \cite{brinkman2010modeling} and \cite{arp2015building} provide the methodologies of building and modeling knowledge graph or medical ontology. \cite{carletti2016ontology} has proposed an implementation of consultation system based on ontology technologies. \cite{dissanayake2020using} reviews clinical decision support system based on clinical ontology reasoning. Inspiration of reasoning over knowledge graph in consultation system can be drew from those works.

Our proposal can be regarded as a conversational AI application in which many machine learning methodologies can be applied, such as reinforcement learning for dialog management and natural language processing for text mining, text information extraction, natural language understanding and comprehension etc. Intuitively, a conversational AI application can be mainly regarded a dialog management with natural language or spoken interface. The core is conversation management since consultation is a dynamic process. Some researches present the method of dialog management based on state tracking. \cite{henderson2014word} propose a word-based dialog state tracking by using Recurrent Neural Network, and \cite{henderson2014second} raise a point of second challenge of state tracking. Xxx represent a reinforcement learning-based spoken dialog system in which the annotation costs are reduced. In 2017, many end-to-end proposals for dialog system emerge rather than hand-craft features based on reinforcement learning and natural language processing methods\cite{wen2016network,bordes2016learning,williams2017hybrid,li2017end,dhingra2016towards}. End-to-end method can be built based on data purely, but the interpretability of the model need to be concerned. In addition, building a medical dialog management system based on natural language processing is a challenge because it needs high quality and domain-specific conversation data. Moreover, making clinical decision needs a serious of complex reasoning process based on medical knowledge, and that’s what common conversation data lacks of, because doctors are less likely to write down the reasoning process on electronic medical records. That’s one of the reasons why we consider using knowledge graph as the core of our proposal. We have simplified natural language interface of our proposal and it will be described in the following parts.

\section*{Knowledge Graph}
\label{sec:headings}
Knowledge graph\cite{stokman1988structuring} is a data structure used to organize knowledge. We build our knowledge graph based on data from SNOMED CT\cite{donnelly2006snomed}. It is composed of multiple facts, each of which consists of a head entity node, a tail entity node and the relationship from the head to the tail. It can also be called a triplet. In our graph, we have 1515 entities, which are divided into two types, disease and symptom. There 574 diseases and 941 symptoms among them. And there are 4065 relations in the graph, all of which are directed, and there are only 1 type of relationship, which is from symptom to disease. The medical knowledge graph is the foundation of the entire intelligent consultation system. The deep learning models, reinforcement learning models, and reasoning algorithms going to be mentioned later are all based on it.

\section*{Methodology}
\label{sec:headings}
There are four core models in the intelligent consultation system, including the translation-based embedding method\cite{bordes2013translating} that maps the entity nodes and relationships of the knowledge graph to the vector space, the diagnosis network responsible for diagnosis, the decision-making network responsible for judging the adequacy of evidence, and the action network responsible for collecting evidence by asking questions related to some certain symptoms.

\subsection*{Translation-based embedding}
Translation-based embedding is a general methodology for a type of knowledge graph embedding technology. The purpose is mainly to map the discrete symbolic concepts from the knowledge graph, including entity nodes to a certain dimension of vector space directly, in order to facilitate the use of vector-based calculation in down stream tasks. We use TransE to embed the basic knowledge graph, but theoretically, all kinds of knowledge graph embedding techniques could be used as well. The objective function of TransE is as follows:
\begin{equation*}
    L = \sum_{(h,r,t) \in S} \sum_{(h',r,t') \in S'} \max(0, f(h,r,t) + \gamma - f(h',r,t'))
\end{equation*}
In which, f is the function for measuring the distance between projected head entity and the tail entity. Usually, the projection is computed by adding head entity and relation in vector space. In our case, L2-norm is taken.
\begin{equation*}
    f = \Vert h + r - t \Vert 
\end{equation*}
In addition, S stands for positive samples, which means actual triplets. And S’ is negative samples , usually they are constructed by alternating elements in original positive samples. The main purpose of the loss function shown above is minimize the distance of real triplets, while maximize the distance of corrupted samples in order to make the model be capable of distinguishing those two kinds of samples.  
During the process of training, first of all, we specify initial values to each vector of corresponding entity:
\begin{equation*}
    e \leftarrow uniform(-\frac{\sqrt{k}}{6}, \frac{\sqrt{k}}{6})
\end{equation*}
then for relationships:
\begin{equation*}
    r \leftarrow uniform(-\frac{\sqrt{k}}{6}, \frac{\sqrt{k}}{6})
\end{equation*}
After that, normalization is taken on all relationships.
\begin{equation*}
r \leftarrow \frac{r}{\Vert r \Vert}
\end{equation*}
On those settings, we start to train the model. In which, k stands for the dimension of entities and relationships. In our case, we take k=512.

The training process adopts 1000 rounds of training strategy, in each round, 500 data are randomly sampled from the set of real triplets as positive samples. The negative sample is formed by randomly selecting 500 positive samples, and then randomly replacing the tail entity nodes with other entity nodes. These data are used as samples and input into the f function for calculation, and finally summarized into the objective function to calculate the gradient, and back-propagated to the vector representing the entity nodes and relationships in a gradient descent manner. After 1000 rounds of training, the model basically converges.

The training process of TransE is used as the pre-training process of the following model, and the obtained entity nodes and relationship vectors will be used in downstream tasks. This process is called transfer learning, and the results obtained from other tasks are used in the current task, so as to avoid the problem of lacking of samples.

\subsection*{Diagnosis Network}
The diagnosis network is a fully connected neural network based on the entity node representing the disease in the knowledge graph and other entities directly or indirectly connected to the current disease node to make a simulation diagnosis. This neural network adopts a three-layer fully connected form, and the dimension of the input layer is consistent with the vector dimension of the entities and relations trained in the previous stage of TransE, which are both 512 in our case. The dimension of the intermediate hidden layer is set to 256, and the output of the hidden layer will be followed by the ReLu activation function, which can avoid the excessive saturation of activation function and obtain more effective back propagation signal. Finally, the dimension of the output layer is consistent with the number of disease type entity nodes in the knowledge graph, and is connected to the softmax function. The purpose of it is to compress the output value to between 0 and 1 and then representing probabilities.

We take Cross Entropy as our objective function:
\begin{equation*}
\begin{split}
L &= -\sum_{n}^{N} y_n \log s_n \\
S_i &= \frac{e^{a_i}}{\sum_{n=1}^{N} e^{a_n}} (i,n=1...N) 
\end{split}
\end{equation*}
Cross entropy is a calculation method often used as the objective loss function for machine learning classification problems. Among them, M is the number of physical nodes of the disease type, and n is the subscript of the current classification, which is the value of the weight obtained by the current classification in the overall proportion, that is, the normalized probabilities.

The training process carries out 1000 epochs for this objective function. In each round, 500 disease entity nodes are randomly sampled from the knowledge graph and search for other entity nodes directly or indirectly connected to them. The search depth of this process is 2. In order to add a random effect to the model to be closer to the real-world situation, a few nodes is randomly dropped. It’s inspired by the dropout regularization method proposed in \cite{hinton2012improving} and leveraged in \cite{krizhevsky2012imagenet} for introducing noise to make model more generalized. Then according to the ID of the node, the corresponding vector is found from the vector lookup table generated by TransE. The relation vector is also found in this way. For the indirectly connected entity vector, the virtual entity vector is obtained by adding the entity vector and the corresponding relationship vector, and then the calculation is performed in the next step; for the directly connected entity vector, no additional operations are required. After obtaining the direct entity vector or virtual vector, the input of the diagnostic network can be obtained by summing these vectors and the forward calculation process can be started. The loss value and gradient are calculated through the above objective function then the gradient descent is performed. Finally the whole training process is terminated until the model converges.

\subsection*{Decision-making Network}
The decision-making network is a model used to judge whether the current evidence obtained is sufficient during the consultation process. If the current evidence is sufficient, the diagnosis network of the previous stage can be used for making diagnosis. This module is very important in the entire model and is the key to determine whether the consultation process can be completed quickly and accurately. This neural network uses a three-layer fully connected neural network form, and the dimensions of the input layer are consistent with the vector dimensions of the entities and relationships trained by TransE, which are both 512. The dimension of the intermediate hidden layer is set to 256, and the output of the hidden layer will be followed by the ReLu activation function as the setting of Diagnosis Network. Finally, the dimension of the output layer is 1, and the sigmoid function is taken in the end. The purpose is to compress the output value to between 0 and 1 and then perform logistic regression.

We take Cross Entropy as the objective function:
\begin{equation*}
L=\left\{
             \begin{array}{lr}
             -\log h_\theta(x), & if \ y = 1 \\
             -\log (1 - h_\theta(x)) & if \ y = 0 \\ 
             \end{array}
\right.
\end{equation*}
In which, $h_\theta$ sigmoid function, x is the output of the last layer and y is the actual result.

Basically, the way of processing nodes or vectors is the same as the one taken in the Diagnosis Network. We set 1000 epochs in the training process. It should be noted here that in the training process, the decision network depends on the diagnostic network generated in the previous stage. When the same data is input into the diagnosis network, if the result of the diagnosis is the disease of the current sample, then this input is a positive sample, and the regression target of the decision network should be 1, indicating that the current obtained evidence is sufficient to make a decision ; If the diagnosis result is not the disease of the current sample, then this input is a negative sample, and the regression target of the decision network should be 0, indicating that the evidence currently obtained is not sufficient and no decision can be made at this time. In other words, the process of consultation should be continued. The loss value and gradient are calculated through the above objective function then the gradient descent is performed. Finally the whole training process is terminated until the model converges.

\subsection*{Action Network}
Action Network is the main core of the entire intelligent consultation system. It’s main responsibility is reasoning the next related symptom and asking related question in the following steps according to the current evidence collected from patients during the process of consultation. This problem is more complex than the above two networks, because the whole process is not static but dynamic, which means we couldn’t simply see this problem as a regular supervised learning process. In practical scenario, users or patients would like a short and accurate consultation process. In other words, an accurate set of diagnosis is demanded and the questions going to be asked needed to be as less as possible. It’s a trade-off problem between these two goals, the accurate diagnosis and less questions. If there are totally 1000 symptoms in the world, we ask patients whether they have those symptoms one by one, then finally we would have a set of strong evidence to infer the disease that the patient get, but it’s impractical because no one can bear such a long conversation process. On the other hand, if we just ask one question related to a certain symptom then the model make inference immediately, then we have a very short process of conversation but there must be very poor reasoning result eventually because the evidence is far from sufficient. Therefore, finding a reasonable question-asking strategy is the main problem. We model it with reinforcement learning problem. Specifically, we use Policy Gradient\cite{sutton1999policy}.

Policy gradient is a classic reinforcement learning method. Mainly, it’s used for searching more strategies to finish tasks in order to get higher grade. In our case, we use neural network to model the actor who interacts with the environment. During training and testing process, we simulate the virtual environment by sampling on the Knowledge Graph. Then Actor takes actions to get observations and rewards in order to learn and evolve. The main setting of the reinforcement learning from different aspects will be shown in the following part.

\subsubsection*{Actor}
Mainly, actor is the main virtual robot doctor we want to train. We model the process of consultation between doctor and patient as the interaction between actor and the environment. In which, doctor takes actions in the form of asking questions about whether the patient have certain symptoms. If the answer is positive, then that means doctor collects evidence in a right direction. If not, it doesn’t mean that it’s useless but an important action for filtering unrelated symptoms and diseases.

We model the actor with a typical neural network which is also in a three-layer fully connected form, and the dimension of the input layer is consistent with the observation returned from the environment. In our case, it is 512 * 3 = 1536. The dimension of the intermediate hidden layer is set to 256, and the output of the hidden layer will be followed by the ReLu activation function. Finally, the dimension of the output layer is consistent with the number of type of symptoms in the Knowledge Graph, in our case, it’s 941. Then it is connected to the softmax function such that the output of this network is a probability distribution over all actions. In real world application, actions will be transformed to some user-friendly questions.

\subsubsection*{Environment}
We form our environment with the Knowledge Graph. We initialize every episode by sampling a disease randomly from it. Then the related symptoms of the disease are obtained from the Knowledge Graph. As what is mentioned above, the dropout method to those symptoms for regularizing is used. We assume that the data could represent typical patients with the disease we sample and the robot doctor will interact with them. 

Observation returned from the environment is composed of three parts. The first part is actions which have already been taken. The second one is related diseases of the taken actions. The last one is other symptoms related to diseases found out in the last part. The three parts are all constructed in the form of vectors. Specifically, in each part, every action already taken will be transformed to embedding vector trained with the translation-based methods and summed up together, then the summation result of each part will be concatenated to be one vector as the observation. Such that, the dimension of the observation going to be input to the actor neural network is 512 * 3 = 1536. 

After obtaining observations from the environment, actions will be taken. Actions that the actor takes are symptoms. As mentioned above, the output of the actor neural network is a distribution over all symptoms, one of those will be chosen as the next action going to be taken. 

If the action the actor takes is in symptoms sampled in the current episode, the environment will return the positive reward of 1. If the actor takes what is already been taken or symptoms not related to the current sample, it gets the negative reward of -1. This setting makes the actor try to find out the critical and important symptoms for making diagnosis.

\subsubsection*{Objective}
The main objective of our task is getting as much reward as possible. In other words, taking correct and effective actions are essential. Such that, the objective function could be modeled as follows:
\begin{equation*}
    R_\theta = \sum_{t=1}^{T} r_t
\end{equation*}
In which, $\theta$ is the policy used by actor to take actions. $r_t$ is the reward returned from the environment in every time step t. And T is the number of total steps in a certain episode. In our case, because actions taken by actor are not deterministic but stochastic, the objective should be transformed to the expectation:
\begin{equation*}
    \bar{R_\theta} = E_{\tau \sim P_\theta(\tau)} \left[ R(\tau) \right] = \sum_{\tau} R(\tau)P_\theta(\tau)
\end{equation*}
In which, $\tau$ is the trajectory which is composed of observations, actions and rewards, usually it’s represented as $\tau = \left\{ o_1, a_1, r_1, ..., o_T, a_T, r_T \right\} $. $o$ is observation, $a$ is action and $r$ is reward. $R(\tau)$ is the total reward actor gets from the trajectory. And $\tau$ is sampled with respect to $P_\theta$. However, the trajectory $\tau$ is endless and the expectation can not be calculated accurately. Such that, the expectation is approximated as follows:
\begin{equation*}
    \bar{R_\theta} = E_{\tau \sim P_\theta(\tau)} \left[ R(\tau) \right] \approx \frac{1}{N} \sum_{n=1}^{N} R(\tau^n)
\end{equation*}
In which, we calculate the approximate expectation by sampling N trajectories from $P_\theta(\tau)$. $P_\theta(\tau)$ can be represented as follows:
\begin{equation*}
    \begin{split}
    P_\theta(\tau) &= p(o_1)p_\theta(a_1|o_1)p(s_2|o_1,a_1)p_\theta(a_2|o_2)p(o_3|o_2,a_2)... \\
    &= p(o_1)\prod_{t=1}^{T}p_\theta(a_t|o_t)p(o_{t+1}|o_t,a_t)
    \end{split}
\end{equation*}
In which, $p(o_1)$ is the probability of the observation $o_1$ and $p(o_{t+1} | o_t, a_t)$ is the probability that observation $o_{t+1}$ is seen after actor takes action $a_t$ on $s_t$. And $\prod_{t=1}^{T}p_\theta(a_t|o_t)$ is the only part which depends on the actor based on $\theta$. It’s the multiplication of probability that actor takes action $a_t$ on observation $o_t$ during the whole trajectory. We model $P_\theta$ with the actor neural network. The gradient calculated in optimization function is as follows:
\begin{equation*}
\begin{split}
\nabla R_\theta(\tau) &= \sum_{\tau}R(\tau)\nabla P_\theta(\tau) \\
&= \sum_\tau R(\tau)P_\theta(\tau) \frac{\nabla P_\theta(\tau)}{P_\theta(\tau)} \\
&= \sum_\tau R(\tau)P_\theta(\tau)\nabla\log P\theta(\tau) \\
&\approx \frac{1}{N} \sum_{n=1}^N R(\tau^n) \nabla \log P_\theta(\tau^n)
\end{split}
\end{equation*}
As what is shown above:
\begin{equation*}
\begin{split}
P_\theta(\tau) &= p(o_1)\prod_{t=1}^{T}p_\theta(a_t|o_t)p(o_{t+1}|o_t,a_t) \\
\log P_\theta(\tau) &= \log P(o_1) + \sum_{t=1}^T \log p_\theta(a_t | o_t) + \sum_{t=1}^T\log p(o_{t+1}|s_t, a_t)
\end{split}
\end{equation*}
In summary, the gradient is as follows:
\begin{equation*}
\begin{split}
\nabla \bar{R_\theta} &\approx \frac{1}{N} \sum_{n=1}^N R(\tau^n) \nabla \log P_\theta(\tau) \\
&= \frac{1}{N} \sum_{n=1}^N R(\tau^n) \sum_{t=1}^{T_n} \nabla \log p_\theta(a_t^n | o_t^n) \\
&= \frac{1}{N} \sum_{n=1}^N \sum_{t=1}^{T_n} R(\tau^n) \nabla \log p_\theta(a_t^n | o_t^n)
\end{split}
\end{equation*}
In which, $p_\theta(a_t^n | o_t^n)$ is the probability that actor takes action $o_t^n$ in a certain episode $n$ and step $t$ after observing $o_t^n$, and $R(\tau^n)$ is the total reward the actor gets during the episode $n$. This expression is intuitive. We hope that the expectation of total reward is as large as possible. Therefore, we need to adjust our policy $\theta$ to make this value larger, and the adjustment is actually to maximize the probability of a certain action which makes the total reward $R(\tau^n)$ finally gotten large. Therefore, we can see that each $\nabla \log p_\theta (a_t^n | o_t^n)$ is weighted and summed with the final reward of the entire episode, instead of multiplying by a certain reward obtained at that time step, because we need to find a strategy to maximize the final total reward in the episode. It should be considered on a global scale, rather than greedily just caring about whether we can get immediate great reward for each action.

\subsubsection*{Parameter Setting and Training Process}
During the training process, we have leveraged some tricks to improve the model performance in terms of efficiency and effectiveness. Firstly, we introduce the reward shaping methods. It’s like a guidance for the actor to learn at the beginning of training process. Simply, we set a reward shaping rate to 1 at the beginning, which means the actor will take actions by following the existing symptoms in the current episode strictly. As the training goes on, the shaping rate decreases gradually with respect to the decreasing rate of 0.9995. For instance, after decreasing 1000 times, the possibility that the actor takes action on its own decision is $1 - 0.9995^{1000} = 39.35\%$. The total number of epochs taken to train actor is 250000, and the reward shaping rate decreases in every 50 epochs. In other words, it has been changed for 5000 times. The curve of decreasing of reward shaping rate is as follows:
\begin{figure}[htp]
    \centering
    \includegraphics[width=8cm]{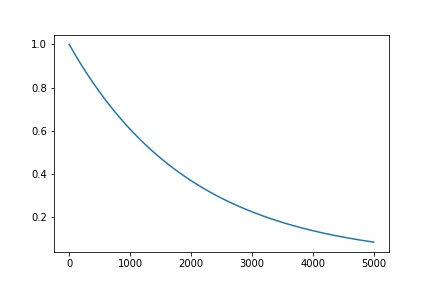}
    \caption{the decay of reward shaping rate}
    \label{fig:galaxy}
\end{figure}
Secondly, we use replay buffer for accelerating training process. Since it takes long time to collect episode data, we decide to leverage the replay buffer to store episode data in advance. Such that, generating data and training can be done in parallel. Specifically, we have a initialized actor $P_\theta$ and it’s copy $\bar{P_\theta}$ which are responsible for training and collecting data respectively. Then actor $\bar{P_\theta}$ firstly interacts with the environment for collecting data, later on the data will be inserted into the replay buffer\cite{lin1992self}. In this process, the model of actor is fixed. Each piece of data includes:
\begin{equation*}
(o_t, a_t, \sum_{i}^{T}r_i)
\end{equation*}
Which are current observation $o_t$, action $a_t$ that the actor took, the total reward $\sum_{i}^{T}r_i$ action gets in the corresponding episode. After the amount of data meets the lowest requirement, it’s 100 in our proposal, batch data, the size is 20, will be sampled from the replay buffer for training actor. After the current iteration training is finished, $\bar{P_\theta}$ will be replaced by the updated $P_\theta$ then continues the collecting process. Moreover, collecting and training are in parallel. Meanwhile, as the process of collecting data goes on, the oldest data will be removed from replay buffer automatically such that the data remains fresh. Finally, in order to increase the randomness in the training process and encourage the actor to find a better path of consultation from a more diversified perspective, we use Boltzmann Exploration\cite{cesa2017boltzmann} method. In other words, the model doesn’t always take the action with max probability but respect to the normalized probability output from the softmax function. That is to test more diverse actions.

\subsubsection*{All in One}
The entire consultation system includes four models, which are TransE model, Diagnosis Network, Decision Making Network and Action Network. TransE model maps nodes and connections among nodes of our knowledge graph into vector space. Those knowledge in the vector space could be directly used in down stream tasks. Action Network collects evidence during consultation process by asking questions related to symptoms or disease reasoned from currently evidence. It needs patients input initial evidence actively at the very beginning, just like the real consultation happens. When patients go to the doctor, usually they will tell the doctor what they feel first. During the evidence-collecting process, Decision Making Network determines whether current evidence is sufficient or not. Diagnosis Network is responsible for making diagnosis based on evidence collected during the consultation process. Once the Decision Making Network determines that the evidence is enough for making diagnosis, the Action Network stops, and the Diagnosis Network makes diagnosis based on evidence immediately.

All those models need to be trained individually, but the order of training need to be aware of. The first one must be translation-based model, in our case, TransE, because other three models must be trained on distributed representation vectors. Secondly, the Diagnosis Network because the training process of Decision Making Network depends on it. Thirdly, the Decision Making Network. While it is being trained, the Diagnosis Network is used to predict disease on evidence, once the predicted one is the same with the one we sample, that is the positive sample. If not ,it is the negative sample. Finally, the Action Network. It depends on all others. We use Decision Making Network and Diagnosis Network to form the virtual environment. Specifically, they are used for computing the reward. 

\subsubsection*{Performance}
After the above-mentioned training process has obtained a set of models with the ability to ask questions and making final diagnosis, a test is performed to evaluate the effect. On the basis of the knowledge graph, 2,000 disease samples are randomly selected and the symptoms or other entity nodes directly or indirectly related to these diseases are randomly removed, and then input into the model for the consultation process and finally a list of diagnosis results is given. The list is output with probability, sorted by probability from highest to bottom, and the results of the highest probability, the top three probabilities, and the top five probabilities are used to calculate the effect. If the target disease appears in the above three lists, it will be taken into account respectively. The overall evaluation results are as follows. It can be seen that in the process of asking symptoms for an average of 15 times, the probability that the final diagnosis result are in the the top 5 diseases reaches 97\%.

\begin{figure}[htp]
    \centering
    \includegraphics[width=12cm]{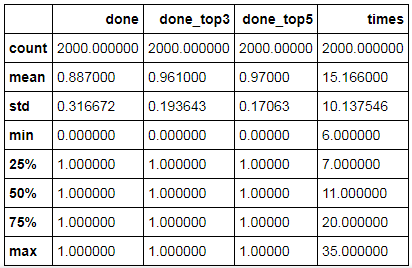}
    \caption{performance}
    \label{fig:galaxy}
\end{figure}

\section*{Discussion}
The reinforcement learning and Knowledge Graph-based consultation system we propose is not perfect solution to address the problem of scheduling the the process of medical consulting. Therefore, some aspects need to be discussed and improved in the future.

\subsection*{The Framework}
What we want to present is a framework for making dynamic decisions continuously, including asking questions and making diagnosis appropriately and efficiently, based on collected information as the process goes on. Hence, different methodologies, such as other translation-based methods or knowledge graph embedding methods, and Q-learning\cite{watkins1992q}, PPO\cite{schulman2017proximal} and other reinforcement learning families could be the substitutes of our methods.

\subsection*{Powered up with NLP}
Users can not seamlessly interact with our consultation system with natural language. Which means that the way of interaction is limited and it results in the obstacle of collecting evidence and making diagnosis effectively and accurately. But from other perspective, leveraging natural language processing might introduce more noise which could make the performance compromised. Therefore, how to use natural language processing to power up our system appropriately should be taken into consideration as well.

\subsection*{Clinical Evaluation}
Our system is evaluated with random sampling data from the Knowledge Graph, in some sense, which is not quite recommended to use in machine learning tasks. Because it’s like use the training samples for testing even though the samples for evaluation is randomly sampled at that stage. For better evaluating our system, we need to test it in clinical trials. By comparing with different doctors or physicians, the real performance can be shown. 

\subsection*{Complementary of Knowledge}
Knowledge Graph is a great way of organizing effective knowledge and by combining with the power of embedding and deep learning techniques, we can reason over Knowledge Graph more easily, effectively and comprehensively. Since our system is built on that, we could not get rid of the problem of knowledge graph completion. Fortunately, there are so many excellent proposals in this topic could be referred\cite{paulheim2017knowledge}. 

Nevertheless, for facilitating the performance of our system during consultation, the knowledge we use should be more specific and detailed. Many other open medical knowledge base, such as UMLS\cite{bodenreider2004unified}, LOINC\cite{mcdonald2003loinc} etc, could be taken into account as well. 

Moreover, since the reasoning rules are not include in our Knowledge Graph, the performance could be improve with the help of effective rules. But how to combine it with embedding and deep learning techniques need to be concerned. Some studies\cite{liang2016neural,callahan2018owl,mohammadhassanzadeh2017semantics} have discussed in this topic.

In addition, data acquisition for updating the Knowledge Graph efficiently worth considering because organizing a large amount of high quality knowledge is a difficult job which consumes a lot of professional manpower and time. Therefore, extracting important knowledge from medical literature and apply it in practice automatically is a critical step to our task.

\subsection*{Practical problem}
Our proposal is simply a prototype. Before applying it to practical scenarios we must consider solutions to the problem of how to cooperate with other systems because it’s impossible to obtain sufficient evidence to support clinical decision making with on-line consultation. Therefore, by depending on off-line service and health state tracking and health management system, our solution can generate value.

%xxxxxx
%xxxxxx

\section*{Conclusion}
We propose a solution to address the problem of the drastically increasing use and short supply of medical resources by leveraging AI technologies. We have described the methodology and framework of building the AI-based consultation system, including the core Knowledge Graph, embedding techniques for mapping entities into vector space, deep neural network for probabilistic reasoning in high dimensional vector space and reinforcement learning manner for discovering more effective and efficient consulting policies. Nevertheless, we have discussed the problems we have encountered but not addressed yet, such as interpretability of the model, clinical evaluation for measuring the performance of our system. In addition, some future works, such as the more user-friendly interface with natural language, complement of the core Knowledge Graph and practical problem we must consider when we apply it in practice, have been discussed as well.

%Bibliography
\bibliographystyle{unsrt}  
\bibliography{references}

\end{document}